\documentclass[11pt]{article}

\usepackage{latexsym,mathrsfs}
\usepackage{amsmath,amssymb}
\usepackage{amsthm,enumerate,verbatim}
\usepackage{amsfonts}
\usepackage{graphicx}
\usepackage{algorithm}
\usepackage{algorithmic}
\usepackage{url}

\newtheorem*{definition*}{Definition}

\newtheorem{remark}{Remark}

\newtheorem*{problem*}{Problem}

\numberwithin{equation}{section}
\numberwithin{table}{section}
\numberwithin{figure}{section}

\DeclareMathOperator{\conv}{conv}

\title{Scalable and Robust Sparse Subspace Clustering  \\ 
Using Randomized Clustering and Multilayer Graphs} 
\date{}

\author{Maryam Abdolali\thanks{Department of Computer Engineering and Information Technology, Amirkabir University of Technology, Tehran, Iran. 
MA acknowledges the support of the Ministry of Science, Research and Technology, Iran. Email: mabdolali@aut.ac.ir} 
\and Nicolas Gillis\thanks{Department of Mathematics and Operational Research,  
Facult\'e Polytechnique, Universit\'e de Mons,  
Rue de Houdain 9, 7000 Mons, Belgium. 
NG acknowledges the support of the ERC (starting grant n$^\text{o}$ 679515) and of the F.R.S.-FNRS (incentive grant for scientific research n$^\text{o}$ F.4501.16). 
Email:  nicolas.gillis@umons.ac.be } 
 \and Mohammad Rahmati$^*$   
}

\begin{document}

\maketitle

\begin{abstract}

Sparse subspace clustering (SSC) is one of the current state-of-the-art methods for partitioning data points into the union of subspaces, with strong theoretical guarantees. However, it is not practical for large data sets as it requires solving a LASSO problem for each data point, where the number of variables in each LASSO problem is the number of data points.  
To improve the scalability of SSC, we propose to 
select a few sets of anchor points using a randomized hierarchical clustering method, and,   
for each set of anchor points, solve the LASSO problems for each data point allowing only anchor points to have a non-zero weight (this reduces drastically the number of variables). 
This generates a multilayer graph where each layer corresponds to a different set of anchor points. 
Using the Grassmann manifold of orthogonal matrices, the shared connectivity among the layers is summarized within a single subspace. 
Finally, we use $k$-means clustering within that subspace to cluster the data points, similarly as done by spectral clustering in SSC.  
We show on both synthetic and real-world data sets that the proposed method 
 not only allows SSC to scale to large-scale data sets, but that it is also much more robust as it  performs significantly better 
 on noisy data and on data with close susbspaces and outliers, 
 while it is not prone to oversegmentation. 


\end{abstract}


\section{Introduction}

Finding a low-dimensional subspace which best represents a set of high-dimensional data points is a fundamental problem in many fields such as machine learning and computer vision. In fact, dimensionality reduction is an essential tool for understanding and preprocessing data sets. 
Using the assumption that the data points have fewer degrees of freedom than the ambient high dimension, 
several methods were developed to discover the underlying low-dimensional structure. 
 Principal component analysis (PCA) is the most popular method for this matter~\cite{jolliffe1986principal}.  

However, the classical methods neglect the fact that the data set often contains data of different
intrinsic structures. For example, facial images of multiple individuals under varying illumination
conditions belong to multiple manifolds/subspaces and should be approximated by multiple
subspaces instead of one~\cite{basri2003lambertian}. 
This leads to a more general problem often referred to as 
\emph{subspace clustering} that has attracted much attention recently. It generalizes the classical PCA in order to
model data belonging to multiple subspaces; see, e.g., the survey \cite{vidal2011subspace} and the references therein. 

Formally the subspace clustering problem can be defined as follows:  
given a set of $N$ data points \mbox{$X = \{ x_i \in \mathbb{R}^d \}_{i=1}^N$} from the union of $n$ unknown subspaces $S_1, S_2, \dots, S_n$ with unknown intrinsic dimensions $d_1,d_2,\dots,d_n$ ($d_i < d$ for all $i$), 
the goal is to partition/cluster the data points according to their underlying subspaces and estimate the subspace parameters corresponding to each cluster. 

Over the past two decades, many methods have been proposed to deal with this problem. They are usually classified in four categories~\cite{vidal2011subspace}: iterative, statistical, algebraic and spectral-based methods. 
Iterative methods \cite{zhang2009median, tseng2000nearest, ho2003clustering} formulate the problem as a non-convex optimization problem and optimize it using a two-step iterative approach, similarly as $k$-means:  Given an initial clustering, alternatively  
(i)~calculate a basis for each subspace, and 
(ii)~assign each point to the closest subspace. 
Despite being simple and intuitive, the convergence is only guaranteed to a local minimum and it is highly sensitive to noise and outliers. 
Statistical approaches~\cite{tipping1999mixtures, sugaya2004geometric} treat the problem as modelling the data with mixture of Gaussian distributions. Similar to iterative methods, statistical methods are sensitive to noise and outliers. 
The algebraic methods~\cite{vidal2005generalized, ma2008estimation} 
fit a set of polynomials to the data points using an algebraic and geometric reformulation of the problem. 
However, not only the methods in this category are sensitive to noise and outliers but also they do not scale well with the increase of the dimension of the data points. 
Spectral-based methods are inspired by classical spectral clustering techniques \cite{von2007tutorial, ng2002spectral}  based on using the spectrum of a specially constructed similarity matrix from the data points. The main difference between methods in this category is in the way the similarity matrix is constructed. 
Global spectral based approaches such as spectral curvature clustering (SCC) \cite{chen2009spectral} tend to construct better similarity matrices (at the expense of a higher computational cost and more sensitivity to noise) compared to local based alternatives \cite{yan2006general, goh2007segmenting}. 
However, with advances in sparse representations (see, e.g., \cite{bruckstein2009sparse} and the references therein), 
a new set of methods has attracted a lot of attention within the spectral-based approaches. 
The key idea is that each data point can be expressed as a sparse linear combination of other data points within the same subspace. A property which is often referred to as \emph{self-expressiveness}.  
Three main representative methods in this category are sparse subspace clustering (SSC) \cite{elhamifar2009sparse, elhamifar2013sparse}, 
low-rank representation based clustering (LRR) 
\cite{liu2013robust, liu2010robust}, 
and least square regression (LSR) \cite{lu2012robust}. 
All three approaches are based on the following model: 
\begin{equation} \label{SSCLRRRLSR}
    \min_{C \in \mathbb{R}^{N \times N}} \; f(C) + \lambda g(E) 
    \quad 
    \text{ such that } 
    \quad
    X = XC + E \text{ and } 
    C_{i,i} = 0 \text{ for all } i, 
\end{equation}
where $X \in \mathbb{R}^{d \times N}$  is the input matrix whose columns are the data points, 
$E \in \mathbb{R}^{d \times N}$ is the noise, 
and $C \in \mathbb{R}^{N \times N}$ is the coefficient matrix. 
 The function $f(.)$ is 
 \begin{itemize}
     \item the component-wise $\ell_1$ norm $||C||_1 = \sum_{i,j} |C_{i,j}|$ for SSC which enhances the sparsity of $C$,  
 \item  the nuclear norm  $||C||_*$ for LRR, that is, the sum of the singular values of $C$, which enhances $C$ to be of low-rank,  and 
 \item  the Frobenius norm $||C||_F = \sum_{i,j} C_{i,j}^2$ for LSR which enhances $C$ to have low energy (this usually does not lead to sparse nor low-rank solutions, but it is computationally much cheaper as the solution can be written in closed form).   
  \end{itemize} 
The function $g$ is a regularization function that is used for modeling the noise 
(usually based on the $\ell_1$ or the Frobenius norm). 
The solution $C$ to \eqref{SSCLRRRLSR} provides crucial information about the links between the data points: $C_{i,j} \neq 0$ means that the data points $i$ and $j$ share some information hence it is likely they belong to the same subspace. These methods then apply spectral clustering on the graph corresponding to the adjacency matrix $|C|+|C|^T$. Note that an important strength of these methods is that they do not need to know the dimensions $d_i$'s of the subspaces, and can estimate the number of clusters as done by spectral clustering (looking at the decay of the eigenvalues of the Laplacian of the adjacency matrix). 

SSC has strong theoretical guarantees in noisy and noiseless cases for both independent and disjoint subspaces \cite{elhamifar2013sparse, soltanolkotabi2012geometric, soltanolkotabi2014robust, you2015geometric}. 
The behavior of LRR in the presence of noise and disjoint subspaces is still not well understood~\cite{wang2013provable}. LSR is  computationally less demanding compared to SSC and LRR, but it is guaranteed to preserve the subspaces only when they are independent. 
Despite theoretical guarantees and empirical success of SSC, the algorithm is not practical for large real-world data sets with more than 10,000 samples, the bottleneck being the resolution of the large-scale convex optimization problem~\eqref{SSCLRRRLSR} (although it can be decoupled into $N$ independent LASSO problems), with a computational cost of $\Omega(N^2)$ operations.  
To overcome this issue, 
we propose a new scalable approach, referred to as \emph{scalable and robust SSC} (SR-SSC), based on solving several problems of the form~\eqref{SSCLRRRLSR} but where only a few rows of $C$ are allowed to be non-zero (using a clever randomized  subsampling of anchor points) in order to generate a multilayer graph that we merge using the technique from~\cite{dong2014clustering}.  
The computational cost is linear in $N$ so that SR-SSC can scale to much larger data sets.

\subsection{Outline and Contribution} 

The main contributions of this paper is to propose a novel scalable approach that extends SSC for clustering large-scale data sets.  
The two key ingredients of this novel method are 
(i)~a randomized hierarchical clustering algorithm that allows to identify different sets of anchor points that are good representatives of the data set, 
and 
(ii)~a multilayer graph technique that summarizes the information across several graphs on the same vertices.  
As we will show, our method significantly improves the performance of SSC in the challenging cases of noisy and close subspaces, and overshadows the oversegmentation issue of SSC.

 The paper is organized as follows. 
 In Section~\ref{relwork}, 
 we briefly review and discuss the related works proposing scalable SSC-like methods. 
In Section~\ref{propalgo}, we present a new approach that makes SSC scalable to large data sets and is more robust, which we refer to as scalable and robust SSC (SR-SSC). 
We investigate the properties and performance of SR-SSC using synthetic and real-world data sets in Section~\ref{numexp}. 
Section~\ref{conc} concludes the paper.

\section{Related Works}  \label{relwork} 

Several methods in the literature have already addressed the scalability of SSC. 
One of the earliest attempts is the method referred to as scalable SSC (SSSC)~\cite{peng2013scalable} that applies SSC on a randomly selected subset of data points and assigns the rest of the data points based on the obtained clusters. 
This approach is very sensitive to the selection method, and suboptimal because the data points which were not selected are not taken into account for generating the clusters. 

In~\cite{you2016scalable}, instead of solving the LASSO problems~\eqref{SSCLRRRLSR}, authors use a greedy algorithm, namely orthogonal matching pursuit (OMP), to obtain the sparse representation for each data point.  
Even though OMP is faster than the original $\ell_1$-based SSC and is scalable for up to 100,000 samples, it is a greedy method with weaker theoretical guarantees~\cite{you2016scalable, dyer2013greedy}. Also, the computational cost is still high, requiring $O(N^2)$ operations.  
In each iteration of OMP to approximate a given data point, the residual is orthogonal to all previously selected data points. 
This property enforces constraints on sampling distribution in each subspace~\cite{dyer2012endogenous}.  
Furthermore, OMP tends to partition subspaces into multiple components (oversegmentation issue)~\cite{chen2018active}. 

Mixture of $\ell_1$ and $\ell_2$ norms  was used in~\cite{panagakis2014elastic} to take advantage of subspace preserving of the $\ell_1$ norm and the dense connectivity of the $\ell_2$ norm. Later in~\cite{you2016oracle}, an oracle-based algorithm, dubbed ORacle Guided Elastic Net solver (ORGEN), was proposed to identify a support set for each sample efficiently. 

More recently, a nearest neighbor filtering approach (KSSC)~\cite{tierney2017efficient} was presented to choose the support set for each sample more efficiently compared to ORGEN. In this approach, the LASSO problem in SSC is restricted to the $K$ nearest neighbors chosen as support set for each point. Even though authors provided some theoretical guarantees for correct connections in noisy and noiseless cases, this approach requires to find the nearest neighbors for each point which requires $O(N^2)$ operations. 
 Additionally this approach is very likely to oversegment the subspaces. 
 Let us illustrate this with a simple example similar to that in~\cite{tierney2017efficient} (we will use a similar example in Section~\ref{numexp}).  
Suppose $8m$  points are chosen around four circles in a 4-dimensional subspace as follows: the columns of the input matrix $X$ are of the form  
\[
 [\cos \theta_k, \sin \theta_k, s \delta, s' \delta]^T, 
\] 
where  $k=0,1,\dots,2m-1$, $\theta_k = \frac{\pi k}{m}$, $s, s' \in \{-1,1\}$ and $\delta \in \mathbb{R}$. 
Let us denote $x_i$ the $i$th column of $X$ ($i = 1,2,\dots,8m$). 
The symmetrized convex hull of the columns of $X$ is defined as 
\[ 
P_X = \conv\left(\pm x_1, \pm x_2, \dots, \pm x_{8m}\right). 
\]  
Solving~\eqref{SSCLRRRLSR} is equivalent to finding the extreme points of the closest face of the polytope $P_X$ for each (normalized) data point~\cite{donoho2005neighborly}.  
It can be proved that by choosing a sufficiently large value for $m$, the vertices corresponding to closest face to the point $[\cos \theta_k, \sin \theta_k, s \delta, s' \delta]^T$ are~\cite{tierney2017efficient}: 
 \[
[\cos \theta_{k\pm 1}, \sin \theta_{k\pm 1}, s \delta, s' \delta]^T, 
 [\cos \theta_k, \sin \theta_k, -s \delta, s' \delta]^T, \text{ and } 
 [\cos \theta_k, \sin \theta_k, s \delta, -s' \delta]^T. 
 \]
Depending of $m$, the value of $\delta$ can be chosen large enough so that the $K$ nearest neighbors are 
$[\cos \theta_{k + \ell}, \sin \theta_{k + \ell}, s \delta, s' \delta]^T$ where $\ell=1,-1,2,-2,\dots$. 
 This leads to the following four disjoint components for the subspaces (4 subspaces of dimension 2), even in the noiseless case:
\[
\left\{ [\cos \theta_k, \sin \theta_k, s \delta, s' \delta]^T \right\}_{k=0}^{2m-1} \quad \text{ with } \quad  s, s' \in \{-1,1\}. 
\] 
This is due to the fact that nearest points do not necessarily contain all the vertices of the closest face in the polytope. More recently,~\cite{traganitis2017sketched} used sketching to speed up SSC, LRR and LSR, but their approach sacrifices accuracy for efficiency.

As a conclusion, we see that there is a need for an efficient algorithm that is computationally cheaper than SSC, 
ideally running in $O(N)$ where $N$ is the number of data points, 
 while not worsening its weaknesses. 
 In the next section, we present such an approach. 
 As we will see, the proposed approach is not only scalable but more robust than SSC and not prone to oversegmentation.

\section{Robust and Scalable Sparse Subspace Clustering Algorithm using Randomized Clustering and Multilayer Graphs}  \label{propalgo}

\begin{figure*}[h!]
\begin{center}
\includegraphics[width=0.8\textwidth]{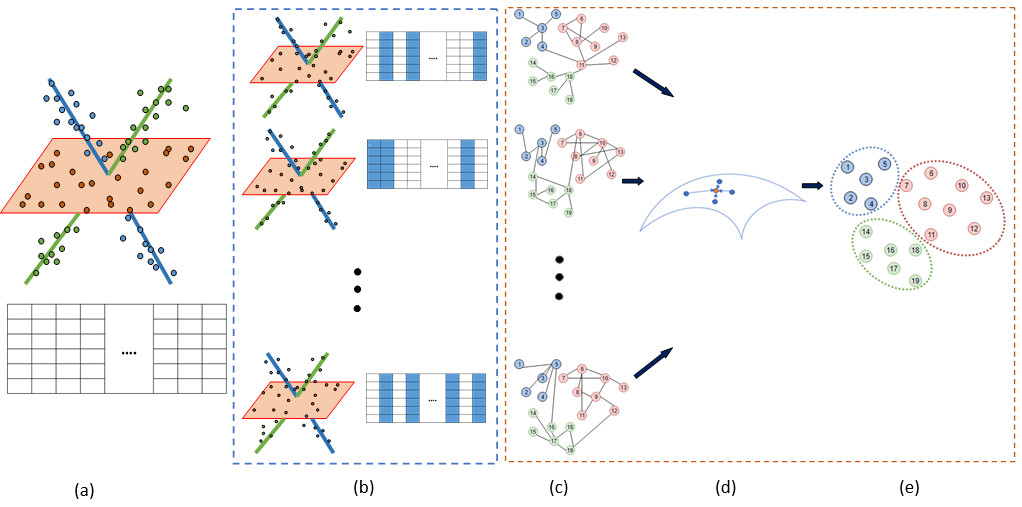}   
\caption{Diagram summarizing scalable and robust SSC (SR-SSC): 
(a) input data points and the corresponding data matrix, 
(b) selection of several sets of anchor points using a randomized hierarchical clustering technique, 
(c) construction of the adjacency graphs corresponding to each set of anchor points by solving~\eqref{SSCLRRRLSR} but only allowing rows of $C$ to be non-zero if they correspond to an anchor point, 
(d) summarization the information among the different graphs using the multilayer graph technique from~\cite{dong2014clustering}, and 
(e) cluster the data points from the final subspace representation using $k$-means. 
}
\label{overal}
\end{center}
\end{figure*}

In order to keep the good performances of SSC while reducing the computational cost, 
we propose an approach in which a small number of samples is chosen as anchor points so that only a few rows of $C$ are allowed to be non-zero in~\eqref{SSCLRRRLSR}, similarly as for the methods presented in Section~\ref{relwork}.  
Clearly, the choice of the anchor points plays a critical role. 
Moreover, the computational cost of the method for choosing the anchor points should be low otherwise there would be no gain in throwing away the rest of the points (in particular, picking the $K$ nearest neighbors is rather expensive; see the discussion in the previous section).  
However, choosing the anchor points with no prior knowledge is a difficult task. 
To reduce the role of the anchor points, we propose a new framework which constructs a multilayer graph based on several sets of well-chosen anchor points using randomization.  
The overall diagram of the proposed framework is shown in Figure~\ref{overal}. 
The details of the two key steps are discussed in the following  two sections, namely the selection of the anchor points and the summarization of the multilayer graph.  

\subsection{Selection of the sets of anchor points via randomized hierarchical clustering} \label{randhier}

In this section, we describe the proposed method for selecting the sets of anchor points. 
Instead of selecting the anchor points fully randomly, we select them in a way so that they are well-spread in the data set, that is, they are good representatives of the data points.  
 To do so, a simple randomized top-down hierarchical clustering technique is used. It works as follows. 
 We start with all the data points in the root node. Then at each step, we 
 \begin{enumerate}
 
 \item Select the node whose sum of the squared Euclidean distances between the data points it contains and their centroid is the largest. 
 Let us denote $I$ the set of indices of the data points corresponding to this node.  
 
 \item Split this node as two child nodes as follows:  
 \begin{enumerate}
    \item Generate randomly a vector $v \in \mathbb{R}^d$ (we used the Gaussian distribution $N(0,1)$ for each entry of $v$), 

	\item  Project each data point $x_i$ within the current node onto the one-dimensional subspace spanned by $v$, using the inner product with $v$, that is, compute, $v^T x_i$ for $i \in I$.  
	
	\item Choose a threshold $\delta$ so that the child nodes 
	\[
	I_1(\delta) = \{ i \ | \ v^T x_i > \delta, i \in I \} 
	 \quad \text{ and } \quad 
	I_2(\delta) = \{ i \ | \ v^T x_i \leq \delta, i \in I \}
	\] 
	satisfy two properties: they are well-balanced (that is, they contain roughly the same number of data points) and stable (that is, modifying $\delta$ slightly does not modify the two child nodes significantly). 
	To do so, we use the technique proposed in~\cite{gillis2015hierarchical} using a simple way to quantify each property:  
	 the fact that the nodes are well-balanced is measured using  $F(\delta) = \frac{|I_1(\delta)|}{|I|}$ which should be close to $\frac{1}{2}$, where $|.|$ indicates the cardinality of a set, and 
	 the stability is measured using 
	 \[
	 G(\delta) = \frac{1}{|I| ( \bar{\delta} - \underline{\delta} )} \left| \left\{ x_i \ | \ \underline{\delta} \leq v^T x_i \leq \bar{\delta}, i \in I \right\} \right|
	 \]
	where $\bar{\delta} = \max(0, \delta - \hat{\delta})$, $\underline{\delta} = \min(1, \delta + \hat{\delta})$, and $\hat{\delta}$ is a small parameter (we used 0.01 as suggested in~\cite{gillis2015hierarchical}).   
To have stable clusters, $G(\delta)$ should be close to zero, that is, modifying $\delta$ should not transfer many data points from one node to the other. 
	Finally, $\delta$ is chosen such that $H(\delta) = -\log\big( F\left(\delta\right) \left(1-F\left(\delta\right)\right) \big)$ + $G(\delta)^2$ is minimized. 
	Of course, many other choices are possible for $H(\delta)$, but the overal algorithm is not too sensitive to this choice as long as both $F$ and $G$ are taken into account. 
	In fact, we have observed for example that choosing $H(\delta) = F(\delta)$ which simply makes the child nodes have the same number of data points (up to one if $|I|$ is odd) 
	does not perform as well. 
    
 \end{enumerate}
  \end{enumerate}
  
The construction of this tree is continued until the number of leaf nodes in the tree reaches the number $k$ of anchor points needed. 
In each node, we select an anchor point which is the data point the closest to the average of the node, that is, to the centroid of the node.  

The computational cost to split a node with $N$ points in dimension $d$ is $O(Nd)$ operations (the most expensive step is to compute the inner products $v^T x_i$ for all $i$). 
To extract $k$ anchors points, the overall procedure hence require at most $O(kNd)$ operations. 
However, unless there are very unbalanced clusters in the data set, 
each child node will contain roughly half the data points of its parent node so that the expected computational cost is actually rather $O(\log(k) N d)$ operations in practice (splitting all the nodes of a single level of the tree requires $O(Nd)$ operations since each level contains all the data points, and there will be $O(\log(k))$ levels in most cases). 

The idea of random projection is, we believe, a key step of SR-SSC because it has the following three key properties that will make it successful: 
\begin{enumerate}

\item It selects anchor points that are good representatives of the data set since they are located nearby centroids of clusters that were generated in such a way that distant points are more likely to be in different clusters than nearby points. 

\item It is computationally efficient, requiring in general $O(\log(k) N d)$ operations to identify $k$ anchor points.  

\item It generates different sets of anchor points, which is essential for our multilayer graph strategy described in the next section. 

\end{enumerate} 

  For example, using random sampling performs worse because it does not satisfy the first property above; see Section~\ref{numexp}.  
Further research includes the design of other approaches to perform the selection of anchor points retaining these three properties.

\begin{remark}
In SSC, (near-)duplicated data points (possibly up to multiplicative factors) is an issue since they tend to be identified as separate clusters. This is one of the reason SSC tends to perform oversegmentation; see also the discussion in Section~\ref{overseg}.   
For example, if $x_i = x_j$ for some $i\neq j$, then we may have $C_{ij}=C_{ji}=1$ (under some mild assumptions on the location of the other data points) while 
$C_{ik} = 0$ (resp.\@ $C_{jk} = 0$) for all $k \neq j$ (resp.\@ $k \neq i$). This will correspond to an isolated vertex in the corresponding graph, which spectral clustering should identify as a single cluster. 
A side result of our randomized anchor point selection step 
is that (near-)duplicated data points are filtered out: 
they will not be selected as anchor points simultaneously (unless the number of anchor points is very large, a situation in which the proposed approach is not very meaningful as it essentially boils down to SSC).  

Another side benefit is that outliers are filtered out as they will not be close to centroids of clusters of points (again, unless the number of anchor points is large in which case an outlier could be identified as a single cluster); see Section~\ref{secout} for some numerical experiments.  
\end{remark}

\subsection{Construction and summarization of the multilayer graph}  \label{multig}

Choosing the number of anchor points depends on many factors: 
the number of subspaces and their dimensions, 
the affinity between subspaces,  
the distribution of data points in each subspace, 
and the level of noise. 
Hence choosing a single set of `good'  anchor points, as done for example in~\cite{peng2013scalable}, is highly 
non-trivial. 
To leverage this difficulty, we propose to choose several sets of different anchor points, as described in the previous section.  
Let us select $L$ sets of $k$ anchor points in the data sets, 
with indices $\Omega^{(j)}$ ($j=1,2,\dots,L$), 
and construct $L$ dictionaries $\{ D^{(i)} = X(:,\Omega^{(i)}) \in \mathbb{R}^{d \times k} \}_{i=1}^L$ (note that we could select a different number of anchor points in each set, but we do not consider this case for simplicity). 
Similarly as for SSC but only allowing anchor points to have a non-zero weight, 
we compute the set of sparse representation coefficients $\{ C^{(i)} \in \mathbb{R}^{k \times N} \}_{i=1}^L$ by solving 
\begin{equation} \label{lassocoef}
\min_{ C^{(i)} \in \mathbb{R}^{k \times N} } || C^{(i)} ||_1 + \frac{\mu}{2} ||X - D^{(i)} C^{(i)} ||_F^2 
\quad \text{ such that } \quad 
C^{(i)}_{j,\Omega^{(i)}(j)} = 0 \text{ for } j = 1,2,\dots,k. 
\end{equation} 

The constraint ensures that no anchor point uses itself for self representation as in SSC.  
The optimization problem~\eqref{lassocoef} can be solved for example using ADMM (Alternating Direction Method of Multipliers)~\cite{boyd2011distributed}; see  Algorithm~\ref{algo1} which we provide for completeness (the superscripts are dropped for convenience). 
ADMM is a good choice for our purpose: it has a low computational cost per iteration (linear in the the number of variables), while its slow convergence (linear at best) is not a bottleneck since a high precision is not necessary as we only need to know the order of magnitude of the entries of $C^{(i)}$ --also, the data is usually rather noisy hence it does not make much sense to solve~\eqref{lassocoef} to high precision.    
 
\algsetup{indent=2em}
\begin{algorithm}[ht!]
\caption{ADMM for~\eqref{lassocoef} for dictionary-based SSC  \label{algo1}}
\begin{algorithmic}[1] 
\REQUIRE 
$X \in \mathbb{R}^{d \times N}$, 
$\Omega$ as the indices of the $k$ column of the dictionary, 
parameters $\lambda$ 
\ENSURE Approximate solution to the problem 
$\min_{C \in \mathbb{R}^{k \times N}} ||C||_1 + \frac{\mu}{2} ||X - X(:,\Omega) C||_F^2$. 
    \medskip  

\STATE Initialization: 
$C=A=\Delta = 0$, $D = X(:,\Omega)$, 
$\mu = \frac{\lambda}{\max_{j, i \neq \Omega(j)} |d_j^T x_i|}$, 
$\rho = \lambda$ 

\WHILE{some convergence criterion is not met }

	\STATE  $A \leftarrow \left( \mu D^T D + \rho I_k \right)^{-1} (\mu D^T X + \rho C - \Delta)$
	
	\STATE  $C \leftarrow T_{\frac{1}{\rho}} \left( A + \Delta/\rho \right)$ \\ where 
	$T_{\gamma} ( y ) = \max \left(0,  |y| - \gamma \right) \text{sign}(y)$ is the soft-thresholding operator 
	
	\STATE  $C_{i, \Omega(i)} = 0$ for $i = 1,2,\dots k$

	\STATE $\Delta = \Delta + \rho (A-C)$ 
	
\ENDWHILE

\end{algorithmic}  
\end{algorithm}

Once the coefficient matrices $\{ C^{(i)} \}_{i=1}^L$ are computed, 
a multilayer graph $G$ with $L$ layers $G^{(i)} = (V,W^{(i)})$ is constructed: 
The set $V$ contains the vertices, one for each data point, and $W^{(i)}$ is the weighted adjacency matrix of the $i$th layer. 
Denoting $E^{(i)}(\Omega(j),:) = C^{(i)}(j,:)$ for $j = 1,2,\dots,k$ 
while the other rows of $E^{(i)}$ are equal to zero, $W^{(i)}$ is obtained by symmetrizing $E^{(i)}$, that is, 
\begin{equation} \label{Widef} 
W^{(i)} = |E^{(i)}| + |{E^{(i)}}^T|, 
\end{equation}
similarly as for SSC. Note that $W^{(i)}$ is a sparse matrix with less than $2kN$ non-zero coefficients. 
Now, we need to combine the information from the individual graphs in each layer. 
To do so, we adopt the method presented in~\cite{dong2014clustering} to merge the information of different layers  into a proper representation such that it strengthens the connectivity/information that majority of graphs tend to agree on. Let us briefly describe this technique. 
The problem of merging a multilayer graph is combined with the problem of merging different subspaces on a Grassmann manifold. 
For each individual graph $G^{(i)}$, we compute its $p$-dimensional subspace representation as the matrix $U^{(i)} \in \mathbb{R}^{N \times p}$ whose columns are the eigenvectors 
corresponding to $p$ smallest eigenvalues of the normalized Laplacian matrix of the corresponding graph, that is,  
\begin{equation} \label{laplace} 
L^{(i)} = I - {D_g^{(i)}}^{-1/2} W^{(i)} {D_g^{(i)}}^{-1/2} 
\end{equation} 
where $D_g^{(i)}$ is the diagonal matrix where each diagonal element is the degree of the corresponding vertex in $G^{(i)}$. 
The problem of combining the multilayer graph is then formulated as follows~\cite{dong2014clustering}: 
\begin{equation} \label{multigraph}
\min_{ U \in \mathbb{R}^{N \times p} } 
\sum_{i=1}^L \text{trace} \left( U^T L^{(i)} U \right) 
	- \alpha \sum_{i=1}^L \text{trace} \left(UU^T U^{(i)} {U^{(i)}}^T \right)
\quad 
\text{ such that } 
\quad
U^T U = I. 
\end{equation}  
The above optimization problem finds a subspace representation that satisfies two goals. 
The first term makes sure the connectivity information of each individual graph is preserved within the subspace $U$ (since this information is contained in their Laplacian matrices $L^{(i)}$).  
The second term is a distance metric that incites the subspace $U$ to be close to the subspaces $U^{(i)}$ corresponding to each graph (on the Grassmannian manifold $U^T U = I$). 
The parameter $\alpha$ balances the two terms. 
The solution $U$ to problem~\eqref{multigraph}  
 is the final subspace representation and can be obtained by finding the $p$ eigenvectors corresponding to the smallest eigenvalues of 
\begin{equation} \label{finallaplace}
L_f = \sum_{i=1}^L  L^{(i)} 
	- \alpha \sum_{i=1}^L U^{(i)} {U^{(i)}}^T.  
\end{equation} 
 The final clusters are obtained by clustering rows of these eigenvectors using $k$-means, as for spectral clustering. As we will see in Section~\ref{alpasec}, the choice of the parameter $\alpha$ will not influence the overal procedure as long as it is chosen in a reasonable range.  
 Note that choosing $\alpha = 0$ amounts to the naive strategy of summing up the adjacency matrices of the different layers, which will turn out to perform rather poorly.

\paragraph{SR-SSC} Finally, our proposed algorithm to perform scalable and robust sparse subspace clustering (SR-SSC) is summarized in Algorithm~\ref{SRSSC}. 
\algsetup{indent=2em}
\begin{algorithm}[ht!]
\caption{ Scalable and Robust Sparse Subspace Clustering (SR-SSC) 
using Randomized Hierarchical Clustering and Multilayer Graphs \label{SRSSC}}
\begin{algorithmic}[1] 
\REQUIRE 
Data matrix $X \in \mathbb{R}^{d \times N}$, 
the number $L$ of sets of anchor points (that is, the number of graphs),  
the number $k$ of anchor points per set, 
the parameter $\alpha$, 
the number of clusters $p$.

\ENSURE A clustering of the columns of $X$ in subspaces  

    \medskip  

\FOR {$i = 1,2,\dots,L$} 

\STATE Choose $k$ anchor points which form the dictionary $D^{(i)}$ 
using the randomized hierarchical procedure described in Section~\ref{randhier}.  

\STATE Solve~\eqref{lassocoef} to obtain $C^{(i)}$ (we use Algorithm~\ref{algo1}). 

\STATE  Construct the symmetrized adjacency matrix $W^{(i)}$ as in~\eqref{Widef}. 

\STATE Compute the normalized Laplacien matrix $L^{(i)}$ as in~\eqref{laplace}. 

\STATE Compute the $p$ eigenvectors $U^{(i)} \in \mathbb{R}^{N \times p}$ corresponding to the smallest eigenvalues of $L^{(i)}$. 

\ENDFOR

	\STATE  Compute the final Laplacian matrix $L_f$ as in~\eqref{finallaplace}. 
	
	\STATE Compute the $p$ eigenvectors $U \in \mathbb{R}^{N \times p}$ corresponding to the smallest eigenvalues of $L_f$. 
	
	\STATE The final clustering of the columns of $X$ are obtained by clustering the rows of $U$ using $k$-means with $p$ clusters.

\end{algorithmic}  
\end{algorithm}

\begin{remark} The choice of  $p$, that is, of the the number of clusters, 
can also be done automatically, as for SSC, for example by looking at the drop in the eigenvalues of the $L^{(i)}$'s and use a majority voting.  
\end{remark} 


\subsection{Computational cost of SR-SSC}   \label{compcost}

Let us analyze the computational cost of Algorithm~\ref{SRSSC}, that is, of SR-SSC. 
Let us first analyze the \texttt{for} loop which is performed $L$ times. 
Identifying the anchor points requires at most $O(kNd)$ operations; see Section~\ref{randhier}. 
The main computational cost for solving~\eqref{lassocoef} using Algorithm~\ref{algo1} is performed at step~3 with a cost of 
$O(k^2N)$ operations to compute $D^T D$, 
$O(k^3)$ operations to compute the inverse, and 
$O(kdN)$ operations to compute $D^TX$. 
Assuming we fix the number of iterations of Algorithm~\ref{algo1} and since $k \ll N$, 
the total computational cost for solving~\eqref{lassocoef} using Algorithm~\ref{algo1} is $O(k^2N + kNd)$. 

Constructing $L^{(i)}$ can be done directly from $C^{(i)}$ and requires $O(kN)$ operations. 
To compute the eigenvectors of $L^{(i)}$, we use ARPACK which is a sparse eigenvalue solver (since $L^{(i)}$ has a most $2kN$ non-zero entries) which requires $O(pkN)$ operations~\cite{Lehoucq1998arpack}. 

It remains to form $L_f$ defined in~\eqref{finallaplace}, and compute the eigenvectors $U$ corresponding to the $p$ smallest eigenvalues. Note there is no need to form matrix $L_f$ explicitly. The first term $\sum_{i=1}^L L^{(i)}$ of the global Laplacian matrix $L_f$ contains at most $O(LkN)$ non-zero elements (since each $L^{(i)}$ has at most $O(kN)$ non-zero elements), and the second term has rank at most $O(L p)$ since each $U^{(i)}$ has $p$ columns. 
Computing the eigenvalue decomposition of a sparse (with $O(LkN)$ non-zero entries) 
plus low-rank matrix (of rank $Lp$) of dimension $N \times N$ using ARPACK requires $O(L N p (k+p) )$ operations. 
In fact, ARPACK is based on implicitly restarted Arnoldi method which reduces to implicitly restarted Lanczos method when the matrix is symmetric (as in our case). Arnoldi/Lanczos are adaptations of the power method for finding a few eigenvalues and the corresponding eigenvectors of a large-scale structured or sparse matrix based on simple iterations using matrix-vector multiplications. Hence it is capable of obtaining the eigenvectors of a matrix for which the explicit stored form is not available as long as the matrix-vector multiplication can be done efficiently. 


The total computational cost of SR-SSC is $O( L N (k^2 + kd + p (k+p) )$ operations,  
which is linear in $N$ hence scalable for large data sets as long as $L$ and $k$ remain small. 
As we will see, a value of $L$ below 10 is usually enough while $k$ should not be significantly larger than the lower bound $\sum_{i=1}^n (d_i+1)$  (since we need to select at least $d_i+1$ points in the $i$th subspace for SSC to work properly). Since the $d_i$'s are usually unknown, a good choice is to pick $k$ as a multiple of $pd'$ where $d' \leq d$ is a guess for the average dimension of the subspaces, e.g., $k = 10 pd'$.  
  
Note that the original SSC algorithm corresponds to tacking $L = 1$ and $k=N$, with a computational cost of $O(N^3 + N^2d)$ operations (for our particular implementation using ADMM). 

\begin{remark}
If the input matrix $X$ is sparse, the computational cost is reduced further as the term $Nd$ is replaced with the number of non-zero entries of $X$. 
\end{remark}

\paragraph{Choice of the parameters $\mu$ and $\lambda$} 

The regularization parameter $\mu$ in~\eqref{lassocoef}, which depends directly on $\lambda$ in Algorithm~\ref{SRSSC}, 
balances the importance between the data fitting term and the sparsity of the coefficient matrix. 
A smaller $\mu$ leads to fewer wrong connections (enforcing sparsity) whereas a larger value for $\mu$ increases the number of true connections (denser connectivity), which  imposes a natural trade-off. 
In other words, we want to set $\mu$ as small as possible while making sure it leads to sufficient connectivity. It was suggested in \cite{elhamifar2013sparse} (and previously in sparse representation literature) to set $\mu$ as $\mu=\lambda\mu_0$ with $\lambda>1$ and $\mu_0 = \frac{1}{\max_{j \neq j} |x_j^T x_i|}$ where $\mu_0$ is the smallest value for which the coefficient vector for some sample would be zero. 
The value for $\mu$ (and hence $\lambda$) that leads to the best results for SSC is data dependent. For data from subspaces with small intrinsic dimension, the regularization parameter should be small to enforce sparsity (as each data point should ideally use $d$ samples to represent itself) and for subspaces with high intrinsic dimension, it should be chosen larger to relax sparsity. Hence, \cite{candes2011robust} suggested the regularization parameter should be in the order of $\sqrt{d}$. 
In this paper, we will fix the value of  $\lambda$ to some prescribed value that leads to reasonable results, 
without giving a particular attention for fine tuning this parameter. 

\section{Numerical Experiments} \label{numexp}

In this section, we evaluate the performance of SR-SSC on both synthetic and real-world data sets. 
All experiments are implemented in Matlab, and run on a computer with Intel(R) Core(TM) i7-3770 CPU, 3.40 GHz, 16 GB. 
The code is available from \url{https://sites.google.com/site/nicolasgillis/code}.

\subsection{Synthetic data sets} \label{secsynth}

In this section, we investigate the performance of SR-SSC under different conditions and depending on the choice of the parameters: the number of graphs $L$, the number of anchor points $k$, and the value of $\alpha$. 
To do so, we use synthetic data sets that consists of three 10-dimensional spaces ($d_1=d_2=d_3=10$) within a 20-dimensional space ($d=20$) with the following bases: 
\[
U_1 \in \mathbb{R}^{20 \times 10} = 
\left( \begin{array}{c} 
\cos(\theta) \, I_{10} \\ 
\sin(\theta) \, I_{10}
\end{array}
\right), 
U_2 \in \mathbb{R}^{20 \times 10} = 
\left( \begin{array}{c} 
\cos(\theta) \, I_{10} \\ 
-\sin(\theta) \, I_{10}
\end{array}
\right), \text{ and } 
U_3 \in \mathbb{R}^{20 \times 10} = 
\left( \begin{array}{c} 
  I_{10} \\ 
  I_{10}
\end{array}
\right), 
\]
where $I_q$ is the identity matrix of dimension $q$ and $\theta \in [0, 2\pi]$. 
For each subspace, we pick $N/3$ random samples (we will choose $N$ as a multiple of 3) generated as linear combinations of $U_i$ where the weights in the linear combinations are chosen at random using the Gaussian distribution of mean 0 and variance 1 (in Matlab, \texttt{$U_1$ * randn(10, N/3)}). 
Finally, an i.i.d.\@ random Gaussian noise with zero mean and standard deviation $\sigma$ is added to the data matrix (in Matlab, \texttt{$\sigma$ randn(d,N)}). Similar to~\cite{soltanolkotabi2012geometric}, the data points are normalized such that their $\ell_2$ norm is equal to one. 

The affinity between subspaces can be defined as an average of the cosine of the angles between subspaces; see~\cite{soltanolkotabi2012geometric}. It was shown in~\cite{soltanolkotabi2012geometric} that the larger the affinity, the more difficult it is for SSC to identify the right subspaces. 
For our particular synthetic data sets, decreasing the value of $\theta$ from $\frac{\pi}{2}$ to 0 increases the affinity and hence makes the subspace clustering task more challenging.  
For evaluating the performance of SR-SSC on this synthetic data set, 
the value for regularization parameter $\mu$ in~\eqref{lassocoef} is chosen as $\mu = \lambda \mu_0$  where $\mu_0$  is the minimum value which avoids a zero coefficient matrix $C$, and $\lambda$ is set as 40.

\paragraph{Budget} 
To have a fair comparison between SR-SSC with different number of graphs and anchor points, 
we define the budget of SR-SSC as 
\[
\text{budget} = L \times k . 
\]
Hence for a fixed budget, a larger vale of $L$ will imply a smaller value of the number of anchor points $k$. Note that this choice of the budget favours SR-SSC with fewer graphs since the computational cost of SR-SSC is not linear in $k$ (it is in $L$); see Section~\ref{compcost}. 
We made this choice for simplicity and to be conservative in the sense that using more graphs will not increase the  computational cost for a fixed budget.  

\paragraph{Accuracy} To assess the quality of a clustering, we will use the accuracy, defined as 
\[
\text{accuracy} = \frac{\text{\# of correctly classified points}}{\text{total \# of points}}. 
\]

\subsubsection{Effect of the parameter $\alpha$} \label{alpasec}

The mutlilayered graph technique from~\cite{dong2014clustering} has a parameter $\alpha$ 
that balances the connectivity of each graph and the distances between their subspace representations; 
see Section~\ref{multig}.  
 The authors in~\cite{dong2014clustering} recommended to use $\alpha = 0.5$ (and observed that their approach is not too sensitive to this parameter). 

\begin{figure*}[h!]
\begin{center}
\includegraphics[width=8cm]{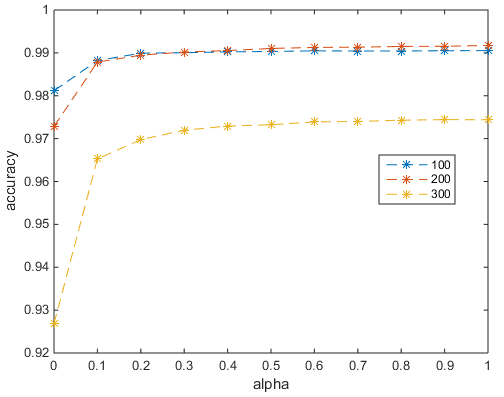} 
\caption{The average accuracy of SR-SSC for different values for the parameter $\alpha$, 
with 100, 200, 300 anchor points. Note that the accuracy decreases for 300 anchor points, this behavior is explained in Section~\ref{numgraan}.} 
\label{alphaplot}
\end{center}
\end{figure*}

To see the effect of $\alpha$ on SR-SSC, we use the synthetic data set described above with $N=3000$, $\sigma = 0.2$ and $\theta = 20^{\circ}$.  
Figure~\ref{alphaplot} displays the average accuracy of SR-SSC with $L = 5$ and $k = 100, 200, 300$ for different values of $\alpha$ over 10 trials. 
We observe that the performance is quite stable for values of $\alpha$ around 0.5. 
(We have made the same observation in our experiments with real data sets although we do not report the results here to limit the size of the paper.) 
It is rather interesting to note that, for $\alpha = 0$, there is a clear drop in the performance of SR-SSC. 
In other words,   
 merging naively the information about the Laplacians $L^{(i)}$'s of each layer 
by summing them together does not work well.

We will use the value of $\alpha = 0.5$ for the remainder of the paper.

\subsubsection{Role of the number of graphs $L$ and anchor points $k$}  \label{numgraan}

In this section, we investigate the performance of SR-SSC depending on the number $L$ of graphs (that is, the number of sets of anchor points) and the number $k$ of anchor points per graph for different affinities between subspaces and noise levels. We will show that selecting multiple sets of anchor points can significantly improve the performance of SR-SSC when the subspace clustering problem gets more challenging 
(that is, larger affinity and higher noise level).

\begin{figure*}[h!]
\begin{center}
\begin{tabular}{c}
(a)~$\theta = 45^{\circ}$ \\ 
\includegraphics[width=7.8cm]{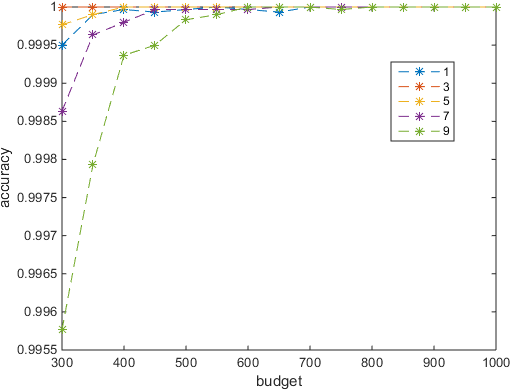} \\  
(b)~$\theta = 30^{\circ}$ \\
\includegraphics[width=7.8cm]{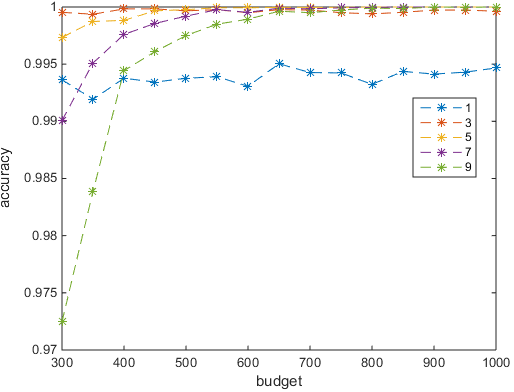} \\ 
(c)~$\theta = 20^{\circ}$ \\ 
 \includegraphics[width=7.8cm]{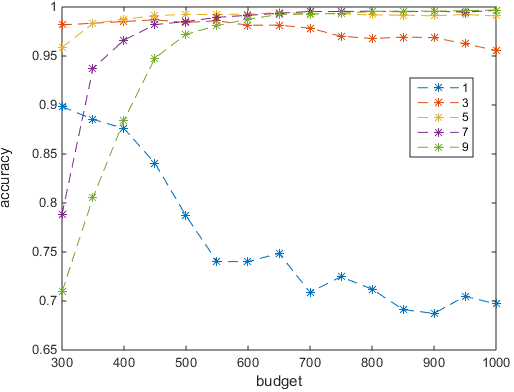}  
\end{tabular}
\caption{Average accuracy of SR-SSC for different number of graphs and for different budgets.} 
\label{numgraph}
\end{center}
\end{figure*}

Let us use the synthetic data sets described in the previous section with $N=3000$ and $\sigma = 0.2$.  
Figure~\ref{numgraph} shows the average accuracy of SR-SSC over 10 generated synthetic data sets 
for different values of the budget, for different numbers of graphs and for 3 values of $\theta$ (45, 30, 20 degrees).  

We observe that if the affinity between subspaces is large enough (namely, for $\theta = 20^{\circ}$), 
the performance of SSR-SSC with one graph decreases as the budget increases; a similar observation was already reported in~\cite{elhamifar2013sparse}.   
 This is due to the fact that as the affinity increases, 
 the chances of choosing wrong connections increases as well. 
 However, the multilayered graph structure reduces 
  this effect as it seeks for connections that are agreed on by a majority of individual graphs.
In fact, we see that for a budget sufficiently large, using more graph improves the performance of SR-SSC significantly. For example, for a budget of 1000, 
SR-SSC with 1 graph (this means that 1/3 of the data points are used as anchors) 
has average accuracy below 70\% while with 9 graphs, it has average accuracy above 99\%. 

\begin{figure*}[h!]
\begin{center}
 \includegraphics[width=10cm]{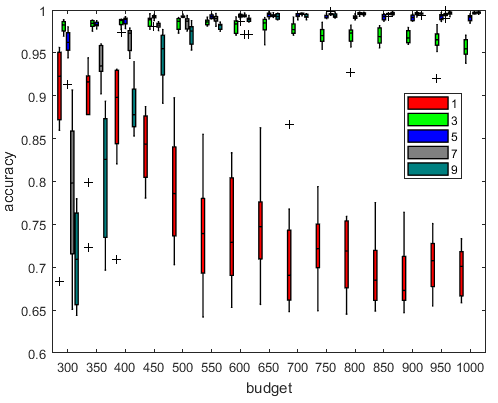}  
\caption{The boxplot of the accuracy over 10 trials for different number of graphs.} 
\label{boxplotgraph}
\end{center}
\end{figure*}

Figure~\ref{boxplotgraph} displays the blox plots for the accuracy of SR-SSC over 10 trails for $\theta = 20^{\circ}$. 
We observe that increasing the number of layers and the number of anchor points not only increases the average accuracy but also reduces the variance significantly.  
Clearly by increasing the number of layers, the chance of choosing good anchor points and hence constructing consistent graphs increases. 
This indicates a natural trade-off between computational cost and increase in performance of SR-SSC. 

\begin{figure*}[h!] 
\begin{center}
 \includegraphics[width=8cm]{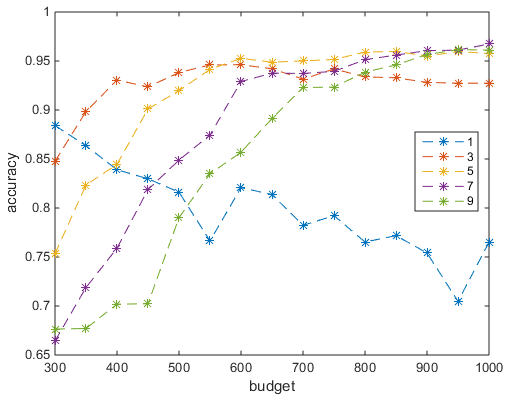}  
\caption{Effect of increasing the noise on the performance of SR-SSC ($\theta = 30^{\circ}$, $\sigma = 0.4$). This figure can be compared with Figure~\ref{numgraph}~(b) 
which corresponds to $\theta = 30^{\circ}$, $\sigma = 0.2$.} 
\label{noisegraph}
\end{center}
\end{figure*}

Noise is the other main factor that makes subspace clustering more challenging. To study the effect of the noise, we use the previous experiment setting with $\theta = 30^{\circ}$  but increase the standard deviation $\sigma$ of the additive Gaussian noise from 0.2 to 0.4. 
The result is plotted in Figure~\ref{noisegraph}. 
Comparing Figure~\ref{numgraph}~(b) and Figure~\ref{noisegraph}, 
it is clear that adding more layers leads to a more stable performance in the presence of a higher noise level. In particular, the average accuracy of SR-SSC with 1 graph drops from over 99\% to below 75\%, while with 5 to 9 graphs, it drops only from 100\% to above 95\%. 

\begin{figure*}[h!] 
\begin{center}
\includegraphics[width=14cm]{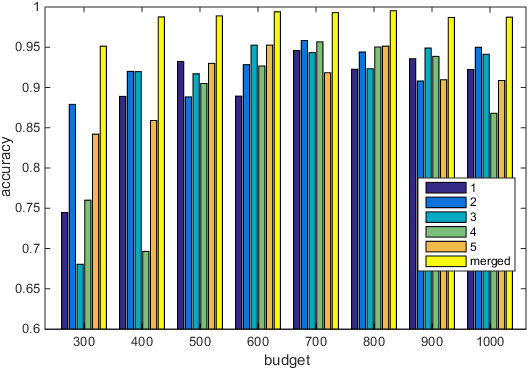} 
\caption{Influence of merging individual graphs on accuracy. The individual performance of 5 layers of graph is compared with the performance of the final merged representation (yellow).} 
\label{merge}
\end{center}
\end{figure*}

It is interesting to compare the performance of individual graphs with the merged one. Let us use the synthetic data sets with $\theta = 20^{\circ}$ and use $L=5$ for SR-SSC. 
The average accuracy with respect to the budget is plotted in Figure~\ref{merge}. We observe that even though the accuracy of individual graphs vary between 70\% to 95\%, 
the performance of the merged graph is stable and around 98\% (for a budget larger than 300). This is due to the fact that the multilayer graph framework is able to highlights the information shared by the majority of graphs. 

\subsubsection{The role of randomized hierarchical clustering for selecting the anchor points} 

\begin{figure*}[h!] 
\begin{center}
\includegraphics[width=8cm]{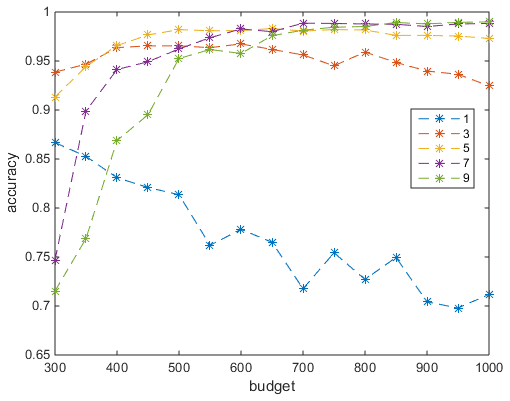}  
\caption{Performance of SR-SSC with a purely random selection of the anchor points. This figure can be compared with Figure~\ref{numgraph}~(c) where the selection is made with our hierarchical clustering technique.} 
\label{hierclus}
\end{center}
\end{figure*}

It is interesting to compare the performance of the randomized hierarchical clustering that is able to identify well-spread anchor points against picking the anchor points uniformly at random.  
Figure~\ref{hierclus} shows the average accuracy over 10 trials of the synthetic data sets with $\theta = 20^{\circ}$  
of SR-SSC with a completely random selection of the anchor points. Comparing with Figure~\ref{numgraph}~(c), 
we observe that the performance of the fully randomized selection of the data points is worse. 
For example, for a budget of 300, accuracy goes 
from 87\% to 90\% for 1 graph, 
from 94\% to 98\% for 3 graphs, 
from 91\% to 96\% for 5 graphs, and 
from 75\% to 79\% for 7 graphs, 
while for 9 graphs it is stable. 
For a budget of 1000, accuracy goes 
from 92.5\% to 95.5\% for 3 graphs, 
from 97.5\% to 99\% for 5 graphs, 
from 99\% to 99.6\% for 7 and 9 graphs, 
while for 1 graph it is stable. 


\subsubsection{Effect of adding outliers} \label{secout}

To analyze the behavior of SR-SSC in the presence of outliers, the same synthetic data set with $N=3000$ under 3 different affinities ($\theta = 45^{\circ}$, $\theta = 30^{\circ}$ and $\theta = 20^{\circ}$) is used. 
We add different numbers of outliers (expressed as a percentage of the 3000 data points) to the data set. 
Each outlier is generated as follows: each entry is generated using the Gaussian distribution of mean 0 and standard deviation 1, and then the outlier is normalized to have unit $\ell_2$ norm as done in~\cite{soltanolkotabi2012geometric}: 
    `This guarantees that outlier detection is not trivially accomplished by exploiting differences in the norms between inliers and outliers'.  
We set the value of the budget to 1000 and evaluate the average performance of SR-SSC in correctly clustering the inliers for different layers of graphs over 20 trials (we apply SR-SSC with 3 clusters on the data set with the outliers, and then compute the accuracy taking into account only the inliers).  

\begin{figure*}[h!] 
\begin{center}
\includegraphics[width=8cm]{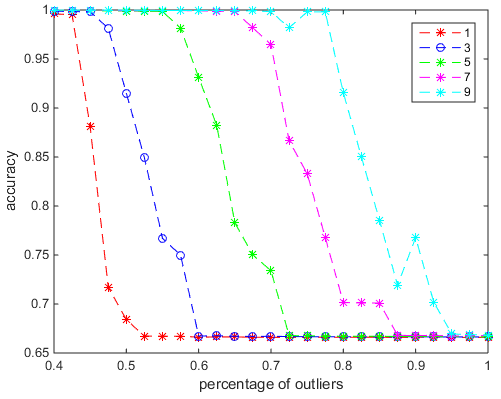} 
\caption{The effect of outliers on performance of SR-SSC for $\theta = 30^{\circ}$, and a fixed budget of 1000. 
The figure reports the average accuracy over 20 runs of SR-SSC for different percentages of added outliers. } 
\label{outlierfig}
\end{center}
\end{figure*}

Surprisingly, for $\theta = 45^{\circ}$, the accuracy remains 100\% for any number of graphs when adding as many outliers as data points. 
For the two other cases ($\theta = 20^{\circ}, 30^{\circ}$), the percentage of outliers which breaks down the accuracy of SR-SSC (with no outliers) is reported in Table~\ref{taboutlier} and the average performance of SR-SSC for $\theta = 30^{\circ}$ over 20 trials is illustrated in Figure~\ref{outlierfig}. 
Not surprisingly, we observe that as the affinity increases, the percentage of allowed outliers to maintain the accuracy decreases. 
Moreover, we observe that adding more layers to the graph (for a fixed budget) leads to a more robust and stable performance as the the percentage of allowed outliers to maintain the accuracy increases significantly. (Recall that that for $\theta = 20^{\circ}$, the one-layered graph performs poorly even without outliers; see Section~\ref{numgraan}.)  

\begin{center}
\begin{table}[h!]
\begin{center}
\caption{Percentage of outliers corresponding to the break-down point of SR-SSC (that is, the accuracy goes below 95\% beyond that point) for different affinities and various number of graph layers. }
\label{taboutlier}  
\begin{tabular}{|c|c|c|c|c|c|c|c|}
\hline 
    $\theta$  & 1 Graph &  3 Graphs & 5 Graphs & 7 Graphs & 9 Graphs \\ \hline
$30^{\circ}$ & 42.5 & 47.5 & 57.5 & 70 & 77.5 \\ 
$20^{\circ}$ & 0 & 2.5 & 17.5 & 20 & 22.5 \\ 
\hline
\end{tabular} 
\end{center}
\end{table}
\end{center}

\subsubsection{Overshadowing the oversegmentation issue} \label{overseg}

\begin{figure*}[h!] 
\begin{center}
\begin{tabular}{ccc} 
\includegraphics[width=6cm]{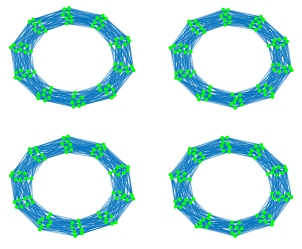} & &  \includegraphics[width=6cm]{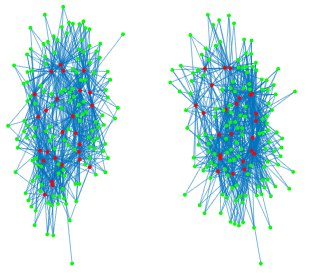} \\ 
(a) & \qquad &  (b) \\ 
\end{tabular}
\caption{Illustration of the oversegmentation of SSC for two subspaces. 
(a) The graph corresponding to the coefficient matrix of standard SSC  with four connected components.  
(b) The graph corresponding to the coefficient matrix of SR-SSC with two connected components. 
The red dots indicate the anchor points. } 
\label{oversegfig}
\end{center}
\end{figure*}

An important issue in SSC is the connectivity of the graphs corresponding to the data points within each subspace.   
In particular, there is no guarantee that for dimensions higher than four, 
the points from the same subspace form a single connected component~\cite{nasihatkon2011graph}. 
In this section, we show that by choosing anchor points as representative samples, 
the oversegmentation issue of SSC is alleviated.   
For this purpose, the points from two subspaces in a 8-dimensional space are created similarly as in~\cite{nasihatkon2011graph}. 
More precisely, we consider 160 points from each subspace chosen around two orthogonal circles. 
Half of the points on the first subspace are given by 
\begin{equation} \label{X11}
x_1(k,s,s') = [\cos \theta_k, \sin \theta_k, s \delta, s' \delta, 0, 0, 0, 0]^T, 
\end{equation}
and of the other half by
\begin{equation} \label{X12}
x_2(k,s,s') = [s \delta, s' \delta, \cos \theta_k, \sin \theta_k,  0, 0, 0, 0]^T, 
\end{equation}
where $\theta = \frac{\pi k}{10}$ ($k=0,1,\dots,19$), 
$s,s' \in \{-1,1\}$ and 
$\delta = 0.1$. 
Similarly, the second subspace contains the points 
\[
y_1(k,s,s') = [ 0, 0, 0, 0, \cos \theta_k, \sin \theta_k, s \delta, s' \delta]^T, 
\]
and 
\[
y_2(k,s,s') = [0, 0, 0, 0, s \delta, s' \delta, \cos \theta_k, \sin \theta_k]^T. 
\]
For a point $x_1(k,s,s')$ defined in~\eqref{X11}, 
the non-zero weights in SSC will correspond to the four points 
$x_1(k\pm 1,s,s')$, $x_1(k\pm 1,-s,s')$ and $x_1(k\pm 1,s,-s')$. 
This implies that there will be no connections between the points $x_1(k,s,s')$ and $x_2(k,s,s')$ within the same subspace, and similarly for $y_1(k,s,s')$ and $y_2(k,s,s')$. 
The graph corresponding to the SSC coefficient matrix $C$ is shown in Figure~\ref{oversegfig}~(a). For this reason, SSC has accuracy of 75\%.  

However, selecting a few anchor points strengthens the connectivity within each subspace. 
The intuition is that if the points use only few anchor points that are well spread within the data set, then the chance of them forming a single connected component in the similarity graph increases. 
This is what we observe when choosing 50 samples (out of 320 data points) on the above data set (using our proposed randomized hierarchical clustering). The graph corresponding to coefficient matrix is shown in Figure~\ref{oversegfig}~(b), and SR-SSC accuracy is 100\%.

\subsection{Real-world data sets} 

In this section, the performance of SR-SSC is evaluated using three large-scale and challenging data sets: handwritten digits (MNIST), object images (CIFAR10) and forest data set (Covertype).  
We compare the accuracy and running times of SR-SSC 
with standard SSC based on ADMM (this is SR-SSC with $L=1$ and $k=N$) 
and three state-of-the art methods for sparse subspace clustering, namely 
OMP~\cite{you2016scalable}, SSSC~\cite{peng2013scalable} (which is closely related to SR-SSC with $L=1$) and ORGEN~\cite{you2016oracle}; see Section~\ref{relwork} for a brief description of these methods.  
For OMP and ORGEN, we used the code available from \url{http://vision.jhu.edu/code/}.

\subsubsection{MNIST data set of handwritten digits} 

The MNIST database contains 70,000 grey scale images of 10 handwritten digits, each of size 28-by-28 pixels.  
For each image, using a scattering convolution network~\cite{bruna2013invariant},  
a feature vector of dimension 3472 is extracted and then projected to dimension 500 by PCA as done in~\cite{you2016oracle,you2016scalable}.  
Nine data sets corresponding to different number of digits are formed (namely digits from 0 to $i$ for $i=1,2,\dots,9$), with around 5000 to 7000 data points of each digit. We also consider data sets with similar digits (2 and 3, 3 and 5, and 1 and 7 are similar), namely \{1,2,3\}, \{1,3,5\}, \{2,3,5\} and \{1,2,3,5,7\}. 
For SR-SSC, we use $L = 5$, and we select $100n$ anchor points where $n$ is the number of clusters. 
For a fair comparison, a dictionary with size corresponding to the whole budget for each case is randomly selected for SSSC. 
The regularization parameter $\mu$ in~\eqref{lassocoef} is chosen as $\mu = \lambda \mu_0$  where $\mu_0$  is the minimum value which avoids a zero coefficient matrix $C$, and $\lambda$ is chosen 120 for our approach, ORGEN and SSSC.   
The sparsity parameter of OMP and regularization parameter of ORGEN are set to 10 and 0.95 according to the corresponding papers~\cite{you2016scalable,you2016oracle}.

The average clustering accuracy and running time for each approach for over 10 trials is reported in Tables~\ref{tabmnist1} and~\ref{tabmnist2}. 
We observe that 
\begin{itemize}

\item SSC runs out of memory for data sets with more than 10,000 points. Since the data set [0:1] has low affinity,, it performs well in this situation. 

\item SR-SSC has the most stable clustering accuracy with a minimum of 89.89\% while being computationally efficient. This validates the fact that the use of a multilayer graph in SR-SSC makes it more robust (given that the number of graphs $L$ and the number of anchor points $k$ is large enough). 

\item OMP performs poorly as soon as the affinity starts to increase. 

\item There is a drop of performance of ORGEN and OMP by adding the digit 3 to the data set.
We investigate this behaviour by plotting the confusion matrix for ORGEN over digits [0:3] in Figure~\ref{oversegfig} (the confusion matrix for OMP is similar). Due to the similarity between the digits 2 and 3, by adding the digit 1, these methods oversegment the cluster containing the digit 1 while failing to distinguish between digits 2 and 3. 
This is confirmed by the low performance of OMP, ORGEN and SSSC in clustering the digits \{1,2,3\} and \{1,3,5\} in Table~\ref{tabmnist1}. 
However, the performance of the aforementioned methods is high in clustering digits \{2,3,5\} which means that unlike  digit 1, the digit 5 cluster forms a well connected component so that digits 2 and 3 are also well separated. 
This confirms our previous results on synthetic data sets in Section~\eqref{secsynth} that showed that our proposed method performs well in the challenging case of close subspaces, and overshadows the oversegmentation effect.   

\item The running time of SR-SSC is similar to the other scalable SSC variants.

\end{itemize}

\begin{center}
\begin{table}[h!]
\begin{center}
\caption{Accuracy (in \%) of the different clustering methods over different sets of digits of the MNIST data set.  
The value M indicates that SSC ran out of 16 GB memory. The best accuracy is indicated in bold, the second best is underlined. }
\label{tabmnist1}  
\begin{tabular}{|c|c|c|c|c|c|c|c|}
\hline 
   Digits   & SSC-ADMM &  OMP & ORGEN & SSSC (500$n$) & SR-SSC (5,100$n$) \\ \hline
$[0:1]$ &99.26 &99.11 &\underline{99.29} &99.21  & \textbf{99.36} \\ 
$[0:2]$ & M  &\textbf{98.67} &98.29 &50.81  & \underline{98.32} \\ 
$[0:3]$ & M &\underline{62.91} &62.67  &62.59 & \textbf{98.06} \\ 
$[0:4]$ & M &69.63 &\underline{69.67}  &64.34 & \textbf{98.23} \\ 
$[0:5]$ & M &48.87 &\underline{75.42}  &65.80 & \textbf{93.41} \\ 
$[0:6]$ & M &76.18 &\underline{78.53}  &68.41 & \textbf{93.98} \\ 
$[0:7]$ & M &69.39 &\underline{80.47}  &76.49 & \textbf{94.26} \\ 
$[0:8]$ & M &59.79 &\underline{81.62}  &75.77 & \textbf{90.53} \\ 
$[0:9]$ & M &48.44 &\textbf{93.85}  &68.07 & \underline{89.89} \\ \hline
$[1\; 2\; 3]$ &M &51.29 &51.10  &\underline{51.37} &\textbf{98.39} \\
$[1\; 3 \; 5]$ &M &52.56  & 52.74 &\underline{55.44} &\textbf{89.20} \\
$[2\; 3 \; 5]$ &M &\underline{95.83}  &\textbf{96.85}  &93.53 &87.96 \\
$[1\; 2\; 3 \;5\; 7]$ &M &64.65  &\underline{70.32}  &61.21 &\textbf{92.58} \\
\hline
\end{tabular} 
\end{center}
\end{table}
\end{center}

\begin{center}
\begin{table}[h!]
\begin{center}
\caption{Running times in minutes of the different clustering methods over different sets of digits on the MNIST data set.  The value M indicates that SSC ran out of 16 GB memory. }
\label{tabmnist2}  
\begin{tabular}{|c|c|c|c|c|c|}
\hline 
   Digits   & SSC-ADMM &  OMP & ORGEN & SSSC (500$n$) & SR-SSC (5,100$n$)\\ \hline
 $[0:1]$ &239.07  &0.69 &2.12 &0.47 &0.64 \\
 $[0:2]$ &M  &1.45 &3.65 &1.08 &1.37 \\
 $[0:3]$ &M &2.36 &5.27 &2.01 &2.48 \\
 $[0:4]$ &M &3.51 &6.96 &3.16 &3.53 \\
 $[0:5]$ &M &5.00 &8.81 &4.66 &5.36 \\
 $[0:6]$ &M &6.37 &10.92 &6.68 &7.56 \\
 $[0:7]$ &M & 8.25 &13.45 &9.03 &11.30 \\
 $[0:8]$ &M &10.85 &15.97 &11.61 &13.73 \\
 $[0:9]$ &M &13.02 &18.28 &14.95 &17.71 \\ 
\hline
\end{tabular} 
\end{center}
\end{table}
\end{center}

\begin{figure*}[h!] 
\begin{center}
\label{confusion}
\includegraphics[width=8cm]{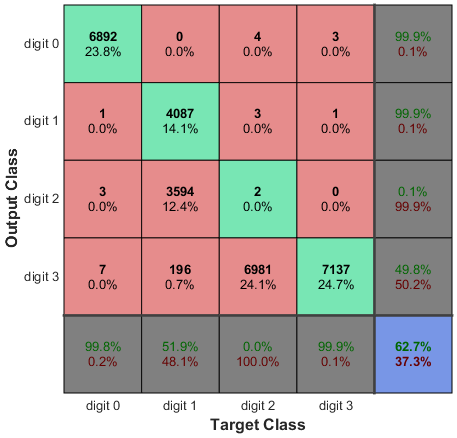}
\caption{Confusion matrix for ORGEN performance over 4 digits of [0:3]. We observe that the digit one is oversegmented, while digits 2 and 3 are clustered together.} 
\end{center}
\end{figure*}

The performance of SR-SSC for different parameters (namely the number of graphs and anchor points) is shown in Tabel~\ref{tabmnist3} for five sets of parameters: (3,30$n$), (5,30$n$), (5,50$n$), (5,100$n$) and (3,200$n$).  
We observe that adding more anchor points increases the accuracy in almost all cases. However, the performance is quite stable among different settings before adding digit 8 and SR-SSC is still able to perform well even with only 30 anchor points per cluster. 
The very similar performances for the (3,30$n$) and (5,30$n$) cases highlight the fact that multilayer framework can improve the performance as long as enough anchor points are chosen for each individual layer. 
Hence the performance of SR-SSC depends on two main factors: 
(i)~to have well-spread and sufficiently many anchor points and 
(ii)~sufficiently many number of layers to best summarize informative shared connectivity among different layers. 
\begin{center}
\begin{table}[h!]
\begin{center}
\caption{Accuracy (in \%) of SR-SSC for different set of parameters over different sets of digits of the MNIST data set.}
\label{tabmnist3}  
\begin{tabular}{|c|c|c|c|c|c|}
\hline 
   Digits   & SR-SSC &  SR-SSC & SR-SSC & SR-SSC & SR-SSC \\ 
   & (3,30$n$) & (5,30$n$) & (5,50$n$) & (5,100$n$) & (3,200$n$) \\ \hline
$[0:1]$ & 97.96 &97.93 &98.81 &99.36  &99.46  \\ 
$[0:2]$ & 96.86 & 97.07 & 97.66 & 98.32 & 98.49 \\ 
$[0:3]$ & 96.47  & 96.81 & 97.73  & 98.06 & 98.54  \\ 
$[0:4]$ &96.72  & 96.61 & 97.74 & 98.23 & 98.51  \\ 
$[0:5]$ &91.68  & 91.54 & 92.46 & 93.41 & 94.02  \\ 
$[0:6]$ &92.12  &92.90 &93.37  &93.98 & 94.02\\ 
$[0:7]$ &92.96  &92.39 &93.36  &94.26 &94.87 \\ 
$[0:8]$ &83.36  &84.01 &86.23  &90.53 &93.95 \\ 
$[0:9]$ &83.95  &82.92 &87.84  &89.89 &91.09  \\ \hline
\end{tabular} 
\end{center}
\end{table}
\end{center}

\subsubsection{CIFAR10 data set of images} 

The performance of the same algorithms is compared on the challenging CIFAR10 data set.  
CIFAR10 consists of 32-by-32 colored images of 10 objects. 
For each image, we converted it to grayscale and then scattering convolution network is used again to extract feature vectors which are projected to dimension 500 using PCA. 
Ten different sets of data corresponding to different numbers of clusters are formed. 
For each cluster, 5000 samples of the corresponding class are selected randomly. We consider 3, 5 and 7 layers of graphs for our method and $100n$ anchor points are selected for each graph. 

The sparsity parameter of OMP algorithm is set to 10 and for the regularization parameter $\mu$ is chosen as for the MNIST data set, with $\lambda = 120$ for all different sets of experiments for ORGEN, SSSC and SR-SSC. The results are reported in Tables~\ref{tabcifar1} and~\ref{tabcifar2}. 

SR-SSC has the highest accuracy in all cases, while it takes less than a minute to cluster 10,000 of data points and less than 15 minutes to cluster 50,000 points. The performance of SR-SSC is slightly higher for 7 graphs compared to 3 and 5 graphs, but the difference is not significant. 

CIFAR is a challenging data set as the linearity assumption of subspaces is violated, which explains the relative low accuracy of SSC and its variants.  However, SR-SSC still manages to outperform all the SSC variants.  
Furthermore, we compared the performances over four extra combinations of clusters at the bottom of Table~\ref{tabcifar1}. There is a clear drop in performance of SSC in clustering [1 2 4] compared to [1 3 4], and [1 2 6] compared to [1 3 6]. This can be explained by the close affinity between custers 1 and 2 compared to 1 and 3. 

In terms of computational time, SR-SSC compares favorably with OMP and SSSC, as for MNIST.  
The computational time of ORGEN is significantly higher compared to other scalable approaches for this data set. 
This is due to the adaptive support set selection method of ORGEN which depends on the correlation between the data points and the \textit{oracle point}. High correlation among data points in CIFAR decreases the amount of discarded data points for each iteration which plays a crucial role for the computational cost of the algorithm.

\begin{center}
\begin{table}[h!]
\begin{center}
\caption{Accuracy of different clustering methods over different sets of clusters on the CIFAR10 data set.  
The value M indicates that SSC ran out of 16 GB memory. The best accuracy is indicated in bold, the second best is underlined. }
\label{tabcifar1}  
\begin{tabular}{|c|c|c|c|c|c|c|c|c|}
\hline 
   \# clusters   & SSC-ADMM &  OMP & ORGEN & SSSC & SR-SSC &SR-SSC & SR-SSC\\
   & & & & (700$n$) & (3,100$n$) & (5,100$n$) &  (7,100$n$) \\ \hline  
2 &	50.68	&50.02  &50.82 &50.01 &80.33 &\underline{82.44} &\textbf{83.15}\\
3 &	33.72	&33.85 &33.89 &	33.45 &\underline{52.01} &51.79 &\textbf{52.86}\\
4 &	M	&25.62  &25.36&	25.06 &51.40 &\textbf{52.29} &\underline{51.76} \\
5 &	M	&20.56	&20.51 &20.07 &39.68 &\textbf{41.71} &\underline{40.97} \\
6 &	M	&17.09	&17.16 &16.70 &\underline{36.58}  &36.56 &\textbf{37.93} \\
7 &	M	&14.59	&14.73 &14.33 &29.83 &\textbf{31.35} &\underline{30.79}\\
8 &	M	&12.79	&12.89	& 12.55	&27.23 &\underline{30.58} &\textbf{31.11}\\
9 &	M	&11.36 &11.51 &	11.16 &26.96 &\underline{27.74} & \textbf{28.07}\\
10 & M&	10.23	&10.37	&10.07	&\underline{24.36} &23.88 &\textbf{26.44}\\ 
\hline
$[1\;2\;4]$ &33.75  &33.36  &33.90 &33.42 &60.69 &\underline{61.85} &\textbf{63.51}  \\
$[1\;3\;4]$ &45.87  &33.39  &33.57 &33.41 &\underline{58.46} &58.25 &\textbf{59.43}  \\
$[1\;2\;6]$ &33.76  &33.93  &33.90 &33.41 &\underline{61.63} &61.20 &\textbf{62.90}  \\
$[1\;3\;6]$ &57.49  &33.35  &33.61 &33.43 &56.08 &\underline{58.39} &\textbf{58.43} \\
 \hline
\end{tabular} 
\end{center}
\end{table}
\end{center}

\begin{center}
\begin{table}[h!]
\begin{center}
\caption{Computational time in minutes of different clustering methods over different sets of clusters on the CIFAR10 data set.  The value M indicates that SSC ran out of 16 GB memory. }
\label{tabcifar2}  
\begin{tabular}{|c|c|c|c|c|c|c|c|}
\hline 
   \# clusters   & SSC-ADMM &  OMP & ORGEN & SSSC & SR-SSC& SR-SSC & SR-SSC \\ & & & & (700$n$) & (3,100$n$) &(5,100$n$) & (7,100$n$) \\ \hline 
2 &	57.69	&0.81 &	6.19	&0.16 &0.18 &0.29	&0.40\\
3 &	162.70	&1.60 &	10.01	&0.40 &0.38	&0.63 & 0.89\\
4 &	M	&2.59	&15.37	&0.80 &0.65 	&1.08 &1.58\\
5 &	M	&3.84	&20.52	&1.35 &1.03	&1.70 &2.38\\
6 &	M	&5.23	&25.36	&2.06 &1.50 &2.59	&3.49\\
7 &	M	&7.08	&29.53	&2.71 &2.22 &3.63 &4.84\\
8 &	M	&8.83	&35.21	&3.85 &3.08	&4.78 &6.54\\
9 &	M	&11.15	&40.27	&5.05 &3.88	&6.07 &8.52\\
10 &M   &13.40 &47.60 &6.12 &4.99 &7.94 &10.74\\
\hline
\end{tabular} 
\end{center}
\end{table}
\end{center}

\subsubsection{Covertype data set} 

The Covertype data set\footnote{http://archive.ics.uci.edu/ml/datasets/Covertype} contains 581,012 samples from 7 categories of tree types. Each sample has 54 categorical/integer attributes that were derived from the data originally obtained from the US Geological Survey (USGS) and USFS data. 
The sparsity parameter of OMP algorithm is set to 15 and $\lambda$ is set to 50 for the regularization parameter $\mu$ for both ORGEN and SR-SSC. 
The results are reported in Tables~\ref{tabcovertype1} and~\ref{tabcovertype2}.

ORGEN achieves the best accuracy, but for a much higher computational cost (see the discussion in the previous section). 
SR-SSC offers the best trade-off between computational time and accuracy. For this data set, the accuracy is not too sensitive to the number of graphs and anchor points. 
Note that SSSC is much faster than SR-SSC in this case because (i) the data set if largen and (ii) SSSC uses SSC on only 4000 samples to learn the subspaces and then cluster the other data points (these are two independent steps). 

\begin{center}
\begin{table}[h!]
\begin{center}
\caption{Accuracy of different clustering methods on Covertype Dataset.}
\label{tabcovertype1}  
\begin{tabular}{|c|c|c|c|c|c|c|c|}
\hline 
    SSC-ADMM &  OMP & ORGEN & SSSC & SR-SSC &SR-SSC &SR-SSC &SR-SSC\\
   & & & (4000) & (5,700) & (3,1400) & (7,700) & (5,1400)\\ \hline  
M	&48.76 &\textbf{53.52} &36.50 &48.35 &48.75 &\underline{48.87} &48.58\\
\hline
\end{tabular} 
\end{center}
\end{table}
\end{center}

\begin{center}
\begin{table}[h!]
\begin{center}
\caption{Computatinal time of different clustering methods on Covertype Dataset (in minutes).}
\label{tabcovertype2}  
\begin{tabular}{|c|c|c|c|c|c|c|c|}
\hline 
    SSC-ADMM &  OMP & ORGEN & SSSC & SR-SSC &SR-SSC &SR-SSC &SR-SSC\\
   & & & (4000) & (5,700) & (3,1400) & (7,700) & (5,1400)\\ \hline  
M	&783 &1452 &14.52 &222.14 &400.12 &323.00 &664.19\\
\hline
\end{tabular} 
\end{center}
\end{table}
\end{center}

\section{Conclusion} \label{conc}

In this paper, we have proposed a new framework to overcome the scalability issue of SSC. Our proposed framework, referred to as scalable and robust SSC (SR-SSC), constructs a multilayer graph by solving LASSO problems using different sets of dictionaries. A fast hierarchical clustering method was used to select the anchor points within the dictionaries that are good representatives of the data set. 
Screening out large numbers of data points drastically reduces the computational cost and memory requirements of SSC. 
Moreover, the multilayer structure has the ability to obtain a summarized subspace representation from different layers of graphs such that it emphasizes shared information among different layers. 

Our experimental results on synthetic and large-scale real-world data sets showed the efficiency of SR-SSC especially in challenging cases of noisy data and close subspaces. 
 Moreover, by choosing few common representative points, the proposed framework has the ability to overshadow the oversegmentation problem of SSC. 
 Of course, there is a trade-off between the computational cost (which is directly related to the number of graphs used and the number of anchor points selected for each graph) and the robustness of SR-SSC, which should be carefully balanced. We have observed in practice that using 5 to 10 graphs is a good choice, while the number of anchor point should be proportional to the number of clusters and their dimensions. 
 
Further work include the improvement of the two key steps of SR-SSC, namely the selection of the sets of anchor points and the summarization of the multilayer graph.  For example, not choosing the different sets of anchor points independently would be particularly interesting in order to make the different layers as complementary as possible. 
Also, studying the theoretical properties of SR-SSC would be a particular promising direction of further research.  
Another direction of research is to improve the implementation of SR-SSC; in particular SR-SSC is especially amenable to parallelization as the constructions of the different layers in the multilayer graph are independent. 


 \paragraph{Acknowledgments} The authors would like to thank Paul Van Dooren for helpful discussion and advice.

\small 

\bibliographystyle{spmpsci} 
\bibliography{SR-SSC}

\end{document}